\documentclass[10pt,twocolumn,letterpaper]{article}

\usepackage{cvpr}
\usepackage{times}
\usepackage{epsfig}
\usepackage{graphicx}
\usepackage{amsmath}
\usepackage{amssymb}
\usepackage{booktabs}
\usepackage{multirow}

\usepackage[pagebackref=true,breaklinks=true,letterpaper=true,colorlinks,bookmarks=false]{hyperref}

\cvprfinalcopy 

\def\cvprPaperID{****} 

\begin{document}

\title{CerfGAN: A Compact, Effective, Robust, and Fast Model for Unsupervised Multi-Domain Image-to-Image Translation}

\author{Xiao Liu\\
Xiamen University\\
{\tt\small xiaoliu95@outlook.com}
\and
Shengchuan Zhang\\
Xiamen University\\
{\tt\small zsc\_2016@xmu.edu.cn}
\and
Hong Liu\\
Xiamen University\\
{\tt\small ynnliu.xmu@gmail.com}
\and
Xin Liu\\
Xiamen University\\
{\tt\small xinliu@stu.xmu.edu.cn}
\and
Cheng Deng\\
Xidian University\\
{\tt\small chdeng.xd@gmail.com}
\and
Rongrong Ji*\\
Xiamen University\\
{\tt\small rrji@xmu.edu.cn}
}

\maketitle

\begin{abstract}
In this paper, we aim at solving the multi-domain image-to-image translation problem with a unified model in an unsupervised manner. The most successful work in this area refers to StarGAN \cite{choi2017stargan}, which works well in tasks like face attribute modulation. However, StarGAN is unable to match multiple translation mappings when encountering general translations with very diverse domain shifts. On the other hand, StarGAN adopts an Encoder-Decoder-Discriminator (EDD) architecture, where the model is time-consuming and unstable to train. To this end, we propose a \textbf{C}ompact, \textbf{e}ffective, \textbf{r}obust, and \textbf{f}ast GAN model, termed CerfGAN, to solve the above problem. In principle, CerfGAN contains a novel component, i.e., a multi-class discriminator (MCD), which gives the model an extremely powerful ability to match multiple translation mappings. To stabilize the training process, MCD also plays a role of the encoder in CerfGAN, which saves a lot of computation and memory costs. We perform extensive experiments to testify the effectiveness of the proposed method. Quantitatively, CerfGAN is demonstrated to handle a serial of image-to-image translation tasks including style transfer, season transfer, face hallucination, etc, where the input images are sampled from diverse domains. The comparisons to several recently proposed approaches demonstrate the superiority and novelty of the proposed method.
\end{abstract}

\section{Introduction}
The image-to-image translation problem aims at translating the input image into the corresponding output image with pixel-level responses \cite{isola2017image}. Followed by the introduction of generative adversarial networks (GAN) \cite{goodfellow2014generative}, extensive research efforts were made. For instance, Isola \emph{et al}. \cite{isola2017image} proposed a general image-to-image translation model in a supervised manner by using conditional generative adversarial nets (CGAN) \cite{mirza2014conditional}. Recent methods like CycleGAN \cite{zhu2017unpaired}, DualGAN \cite{yi2017dualgan} and DiscoGAN \cite{kim2017learning} are proposed to translate images in an unsupervised manner by training two generative networks. However, the aforementioned works focus on two-domain image-to-image translation, where the model only translates one domain to the other. When encountering multiple domains, these approaches are limited and inconvenient, since every pair of image domains requires a model to be built independently.

In terms of multi-domain image-to-image translation, StarGAN \cite{choi2017stargan} serves as a representative work, which uses a single model containing one generator (encoder and decoder) and one discriminator to translate multi-domain facial images in an unsupervised manner. In principle, the classification loss and the adversarial loss in StarGAN force the generated images to fall inside the target domain, which has demonstrated superior performance on tasks like face attribute modification, where all domains are slightly shifted. However, StarGAN performs worse in general translations such as facade labels $\rightleftharpoons$ photos translation \cite{isola2017image} \footnote{A systematic analysis of StarGAN for general image-to-image translation tasks is detailed in Sec. \ref{SecStar}.}. We attribute the poor performance of StarGAN in two aspects: (1) The translated outputs only have a slight change comparing to the initial inputs. (2) The output images share a very similar mode to each other. The main problem is that the classification loss and the adversarial loss can not work well together. To the end, the general multi-domain translation remains unsolved by using a single model.

\begin{figure}[!htb]
\centering
\includegraphics[width=0.45\textwidth]{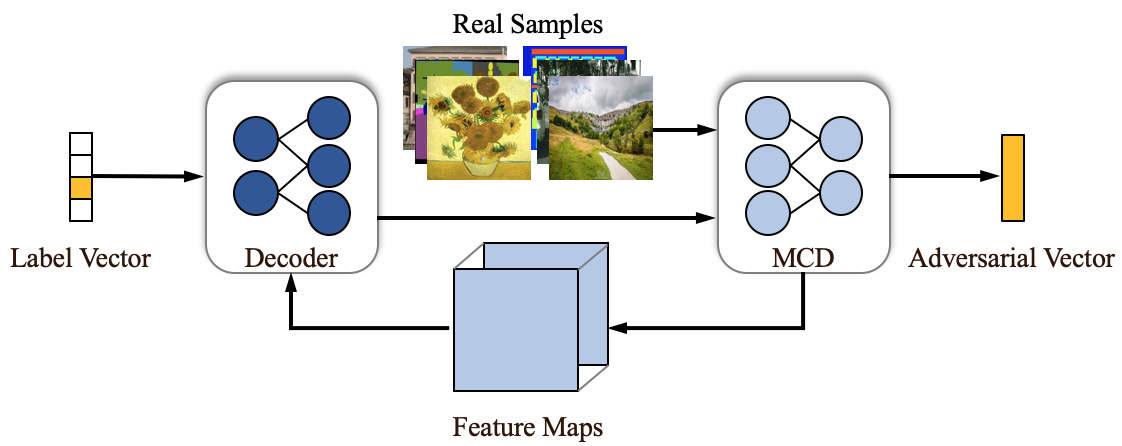}
\caption{Our proposed framework. The output of MCD is a vector rather than a scalar. The decoder translates the feature maps of the MCD and the label information to the target image.}
\label{Fig.Brief}
\end{figure}

To address these problems, we propose a \textbf{C}ompact, \textbf{e}ffective, \textbf{r}obust, and \textbf{f}ast GAN model, termed CerfGAN, which enables accurate and efficient mappings among different domains. To be specific, the real images are classified in multiple domains $X_i, \{i = 1,2,3,\cdots,N-1\}$, where $N$ is the number of domains. In contrast to the discriminator (binary discriminator) in the vanilla GAN \cite{goodfellow2014generative} which outputs a scalar to determine if the input is real or fake, we propose a novel multi-class discriminator (MCD) in CerfGAN, which outputs a $1\times N$ vector $d$. $d(i)$ is the $i^{th}$ element of $d$, which is defined to tell if the input is the real $X_i$. Training the decoder by MCD can also be formed as a new type of conditional GAN \cite{mirza2014conditional}, where the conditional information is used to choose which dimension of $d$ should be output as the adversarial loss. On the other hand, MCD can be simply considered as $N$ binary discriminators sharing most of the weights, where the $i^{th}$ discriminator is responsible to tell if the input is the real $X_i$. In this case, MCD and the decoder in CerfGAN play a multi-player game, which gives powerful \emph{effectiveness} and \emph{robustness} to the unified model. In the experiments, CerfGAN is trained by inputting images sampled from 8 datasets, and state-of-art results are obtained by only using a single model.

CerfGAN also overcomes several drawbacks in the existing
translation architectures. In particular, the cutting-edge pix2pix \cite{isola2017image} used EDD to encode the input image and the conditional information first. Then, the decoder decodes the latent representations of the image to be the target image, where the encoder and the decoder are trained as a generator. The network is much more difficult to train with
gradient descent when the network is deeper due to gradient exploding and vanishing as discussed in\cite{he2016deep} and \cite{bengio1994learning}. In addition, the encoder is a down-sampling network and the decoder is an up-sampling network, which makes the model more unstable to train. To overcome this problem, pix2pix used U-Net \cite{ronneberger2015u} and CycleGAN adopted ResBlock \cite{johnson2016perceptual}. We notice that the encoder in EDD encodes the input image to the latent representation, and the input image also has to be encoded by the discriminator in EDD. So in CerfGAN, we propose to use the feature maps of the MCD as the latent representation of the input image. When training the decoder, we freeze the MCD. When the parameters of the MCD are updated, the decoder is fixed. As a result, only a down-sampling or an up-sampling network will be trained at a time, which makes CerfGAN more stable to train. Meanwhile, as CerfGAN has only two networks, there are much fewer parameters comparing to StarGAN. In our experiments, CerfGAN has 64.1\% parameters of StarGAN and the training time is around 79.17\% of StarGAN, referring to the \emph{compact} and \emph{fast} merits of CerfGAN.

To verify the proposed approach, we perform extensive experiments on many datasets. The state-of-the-art results are achieved by CerfGAN with comparisons to several recently proposed approaches. Fig. \ref{Fig.Brief} illustrates the proposed CerfGAN model.

Our contributions are concluded as follows:
\begin{itemize}
\item We analyze that the classification loss and the adversarial loss of StarGAN are unsuitable for general
image-to-image translation with large domain shifts.
\item We propose MCD to address the problem of StarGAN
and testify the effectiveness of training a decoder by
MCD as a new type of conditional GAN.
\item We propose CerfGAN that utilizes MCD to conduct
encoding, which is compact (\emph{i.e.}, only 64.1\% parameters of StarGAN), effective and robust (\emph{i.e.}, a single model handles 8 datasets with state-of-the-art performance), fast (\emph{i.e.}, around 79.17\% training time of StarGAN).
\end{itemize}

\section{Related Work}
\textbf{Generative Adversarial Networks} \cite{goodfellow2014generative,zhao2016energy} have been widely used in many computer vision tasks including image generation \cite{arjovsky2017wasserstein,huang2017stacked,radford2015unsupervised,zhao2016energy}, super-resolution \cite{ledig2016photo} and image-to-image translation \cite{choi2017stargan,isola2017image,kim2017learning,liu2017unsupervised,yi2017dualgan,zhu2017unpaired}. The discriminator of GAN learns to distinguish whether the input samples are real or fake, referred to binary discriminator in this paper. The generator is trained to translate the inputs to fake images to fool the discriminator. As many computer vision tasks can be considered as translations, it is a common solution to train a generator by giving prepared data and optimizing the adversarial loss. Instead, in our framework, we propose to use the adversarial loss to train the decoder to produce target-domain images as realistic as possible.

\textbf{Conditional GAN} (CGAN) is proposed to generate samples by giving conditional information. As a conditional version of GAN, CGAN simply feeds the data and its corresponding conditional information to both the generator and the discriminator. The discriminator in CGAN is responsible to distinguish if the sample is paired with the conditional information. The generator learns to generate corresponding samples when being given certain conditions. 
\begin{figure}[!htb]
\centering
\includegraphics[width=0.45\textwidth]{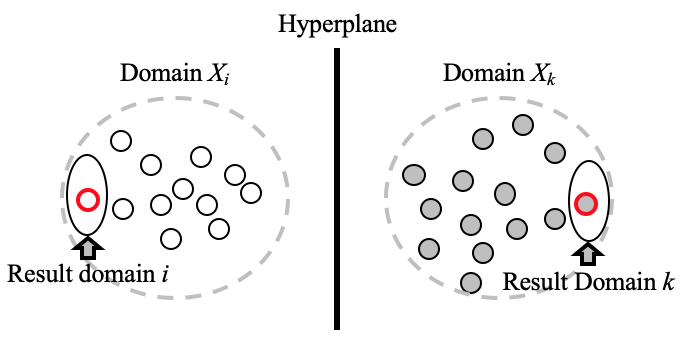}
\caption{Analysis of the minima of the loss. The line in the middle is the hyperplane of the classifier. The results of StarGAN will be in the result domains. More details in Sec. \ref{SecStar}}
\label{FigStarGANDiss}
\end{figure}
In MAD-GAN \cite{ghosh2017multi}, the discriminator outputs a vector rather than a number to distinguish if the sample is real and to identify the generator that the fake sample comes from. MAD-GAN can also be regarded as a conditional version of GAN, which uses the condition to train the discriminator to identify the generators, rather than feeding the conditional information and the data to the discriminator directly. The discriminator of MAD-GAN shares some similarities with the proposed MCD. The difference is that MCD in our framework outputs a vector that distinguishes if the input is from a specific domain or is output by the decoder.

\textbf{Adversarial Auto-Encoder} (AAE) \cite{makhzani2015adversarial} uses the adversarial training to match the latent distribution with the arbitrary prior distribution in the latent space. By training the generator to map the prior distribution to the data distribution, AAE can be applied in many tasks such as dimensionality deduction, clustering and data visualization. The architecture of our framework is similar to AAE. However, the training process and the principle are totally different. The MCD in our framework also serves as the encoder with the main role to match the generated data distribution with the target data distribution. 

\textbf{Image-to-Image Translation} is a research hot spot, as various computer vision tasks can be classified as an image-to-image translation problem, such as super-resolution \cite{ledig2016photo} (low-resolution to high-resolution), style transfer \cite{johnson2016perceptual} (photo $\rightleftharpoons$ artistic style), face hallucination \cite{wang2014comprehensive} (face photos $\rightleftharpoons$ sketches) and so on. pix2pix \cite{isola2017image} introduced framework of conditional adversarial networks for a general-purpose solution to image-to-image translation problem in a supervised manner. To further break the need of paired data in pix2pix, several unsupervised image-to-image translation frameworks are proposed in  \cite{liu2017unsupervised,zhu2017unpaired}. For instance, CycleGAN \cite{zhu2017unpaired}, DualGAN \cite{yi2017dualgan} and DiscoGAN \cite{kim2017learning} train two networks to learn the translation mappings between two domains by utilizing cycle-consistency loss \cite{zhu2017unpaired}. And the cutting-edge StarGAN adopts a single model to learn all the translation mappings among multiple domains, \emph{e.g.}, face attributes. 

\begin{figure}[!htb]
\centering
\includegraphics[width=0.4\textwidth]{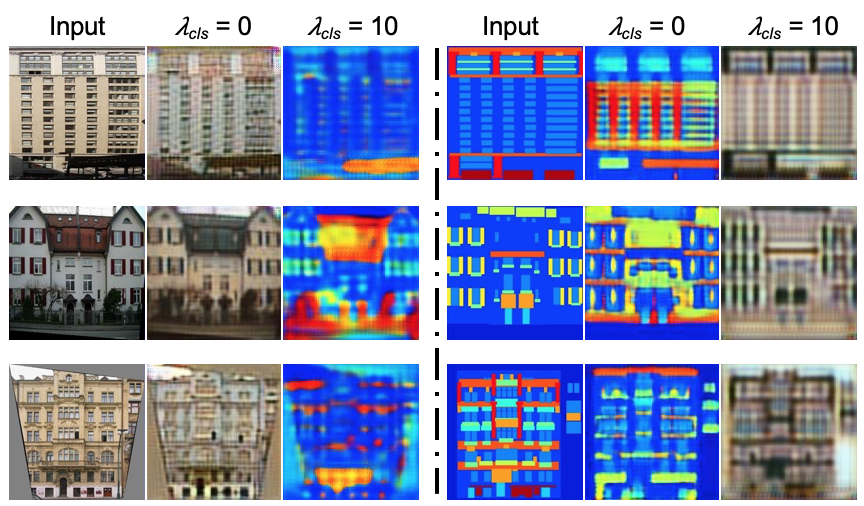}
\caption{StarGAN ($\lambda _{cls} = 0$) and StarGAN ($\lambda _{cls} = 10$) learn to match the translation mappings of face sketch $\rightleftharpoons$ face photo.}
\label{FigStarCls}
\end{figure}

\section{Analysis of StarGAN}
\label{SecStar}
StarGAN utilizes a discriminator and an auxiliary classifier with the cycle-consistency loss (reconstruction loss) \cite{zhu2017unpaired} to train the generator (decoder and encoder). The object functions of StarGAN are: 
\begin{equation}
\label{equStarLD}
    L_D = -L_{adv} + \lambda _{cls}L^r_{cls},
\end{equation}
\begin{equation}
\label{equStarLG}
    L_G = L_{adv} + \lambda _{cls}L^f_{cls} +  \lambda _{rec}L_{rec},
\end{equation}
where $D$ and $G$ denote the discriminator and the generator (decoder and encoder). In Fig. \ref{FigStarGANDiss}, suppose the images of domain $X_i$ are all in the left circle and the images of domain $X_k$ are all in the right circle. If the generated image is in these two circles, the discriminator is fooled by the generator, which means the GAN works very well. In other words, the minima of the adversarial loss are achieved when the output images are in these two circles. 

If there is no classification loss ($\lambda_{cls} = 0$), as shown in Fig. \ref{FigStarCls}, the adversarial loss and the reconstruction loss will only guide the generator to output results that are very close to the initial inputs, which means that the minima of the adversarial loss and the reconstruction loss are achieved when the output images are in the circle where the inputs are in. Thus, the images are not translated to the desired domain. To handle this, StarGAN proposes an auxiliary classifier to force the images to be translated.

If $\lambda _{cls}$ is not 0, principally, the generated images have to be on the proper side of the hyperplane. For example, if the input is from domain $X_i$, the output will be on the right side of the hyperplane, since the class of the output is domain $X_k$ and the classification loss will be very large once the output is in the left side. In this case, the adversarial loss, the reconstruction loss and the classification loss will force the results to be on one side of the hyperplane as well as in the circle on that side. As StarGAN chooses the cross-entropy loss as the classification loss, the loss will be smaller when the result is farther from the hyperplane, which causes the generated image to be away from the hyperplane as much as possible. To sum up, the minima of these three losses have to satisfy three requirements: the output image is on one side of the hyperplane, in the circle and away from the hyperplane as much as possible. As shown in Fig. \ref{FigStarGANDiss}, the results must be in the results domains. 

If the samples in the target domain have similar modes, these samples in the corresponding result domain will contain most of the statistical properties of the target domain. For example, in terms of face attribute modulation, the images in the same domain are similar to each other except a particular attribute like the color of hairs, the size of nose, \emph{etc}. In this case, it is satisfactory to use StarGAN to perform the translation, where slight changes between the inputs and the outputs are required. However, when the samples in the target domain are diverse, the results in the corresponding result domain will only contain a small part of statistical properties of the target domain. Although these output images in the result domains are correctly translated, they are still not satisfactory for general image-to-image translation tasks. It explains why StarGAN fails to solve more complex and general translation tasks like facade labels $\rightleftharpoons$ photos. 

\section{The Proposed CerfGAN}
\label{CerfGAN}
\subsection{Multi-Class Discriminator}
We propose the multi-class discriminator (MCD) to address the problems of StarGAN. In details, MCD does not output a scalar but a $1\times N$ vector when there are $N$ domains. Principally, the image-to-image translation problem is solved by conditional GAN methods. We first model
the decoder training by MCD as a new type of conditional
GAN. Then, we explore how MCD can perfectly solve the
problems of StarGAN.

In terms of CGAN \cite{mirza2014conditional}, the generator (a decoder) is trained by the discriminator that distinguishes if the output and the conditional information are paired, where the input of the discriminator is the conditional information concatenated with the input images. In terms of training the decoder by MCD, the conditional information is not input to MCD directly, but is used to choose the dimension of the adversarial vector. For example, when training MCD with MNIST dataset \cite{lecun1998gradient}, if the desired output is an image of digit 0, the first dimension of the adversarial vector will be output as the adversarial loss, where the number of dimensions $N$ is 10.

Similar to CGAN, the objective function of training a
decoder by MCD can be expressed as:
\begin{equation}
\label{equMCDGAN}
    \begin{aligned}
    \min_{DE} \max_{M} V(M,DE) = \qquad \ \qquad \ \qquad \ \qquad \ \qquad
    \\E_{x^i\sim p_{data}(X_i)}[\log M(x^i|c_i)(i)] \quad + \qquad \ \qquad \\E_{z\sim p_{z}(z)}\Big[\log\Big(1 - M\big(DE(z|c_i)\big)\big(i\big)\Big)\Big], \quad
    \end{aligned}
\end{equation}
where $M$ is the MCD and $M(\cdot)(i)$ is the $i^{th}$ element of the adversarial vector that is responsible to distinguish if the input image is real $X_i$. Note that $DE$ denotes the decoder. For an image generation problem, the prior distribution is denoted as $p_{z}(z)$. $c_i$ is the label of the image that is sampled from the domain $X_i$. MCD is trained by maximizing the objective function in Eq. \ref{equMCDGAN} and the decoder is trained by minimizing it. 

MCD can also be regarded as that $N$ binary discriminators share most of the weights excluding the output layer. In this case, MCD and the decoder do not play a two-player game but a multi-player game. The decoder has to learn to fool all the discriminators combined in MCD, which enables the matching of multiple translation mappings to the model. As a result, CerfGAN is an effective and robust model for translating images sampled from multiple domains,
where the datasets are considerably different. 

To demonstrate how MCD addresses the problems of StarGAN, we suppose the translation mappings are between domain $X_i$ and domain $X_k$ and use MCD to replace the original discriminator in StarGAN. If the input is from domain $X_i$, the $i^{th}$ dimension of the adversarial vector will be output as the adversarial loss. The generated images will only be in the left circle in Fig. \ref{FigStarGANDiss} when MCD and the generator (encoder and decoder) are well trained. In sum, comparing to StarGAN that only generates images in the result domains, training the generator by MCD allows the generator to generate images that contain most statistical properties of the target domain.

\subsection{Encoded by MCD}
\label{SecDD}
The Encoder-Decoder-Discriminator architecture (EDD) proposed in pix2pix \cite{isola2017image} trains the encoder and the decoder simultaneously. Then, these two networks are combined as the generator. This architecture is widely used in lots of subsequent methods \cite{choi2017stargan,ghosh2017multi,hoffman2017cycada,liu2017unsupervised,wang2017high,zhang2017stackgan,zhu2017unpaired,zhu2017toward} in image-to-image translation.
However, in experiments, this architecture
is unstable and hard to train. To address this problem, in CerfGAN, MCD also carries out encoding, where the input
of the decoder is the feature maps of MCD. Comparing to EDD-based models, CerfGAN mainly has two advantages. (1) More stable: EDD-based models train the encoder and the decoder simultaneously to translate the input images. In CerfGAN, the input images are encoded by MCD whose parameters are frozen when training the decoder. As a result, CerfGAN only trains one network at a time, which makes CerfGAN much more stable to train and avoid the gradient vanishing/exploding issue as discussed in He \emph{et al.} \cite{he2016deep}. (2) More efficient: CerfGAN only needs two networks (decoder and MCD) to solve image-to-image translation problems. So CerfGAN has much less parameters, which requires less computation and memory costs in model training and inference. In addition, considering MCD as the simulator of the real world, the feature maps of MCD can be regarded as the real-world rules in the simulator. Intuitively, if the input of the decoder is the feature maps of MCD, the decoder will be easier to train, since the decoder knows more knows more about the actual output distribution, which means it is easier to fool the simulator.

\begin{figure*}[!htb]
\centering
\includegraphics[width=0.9\textwidth]{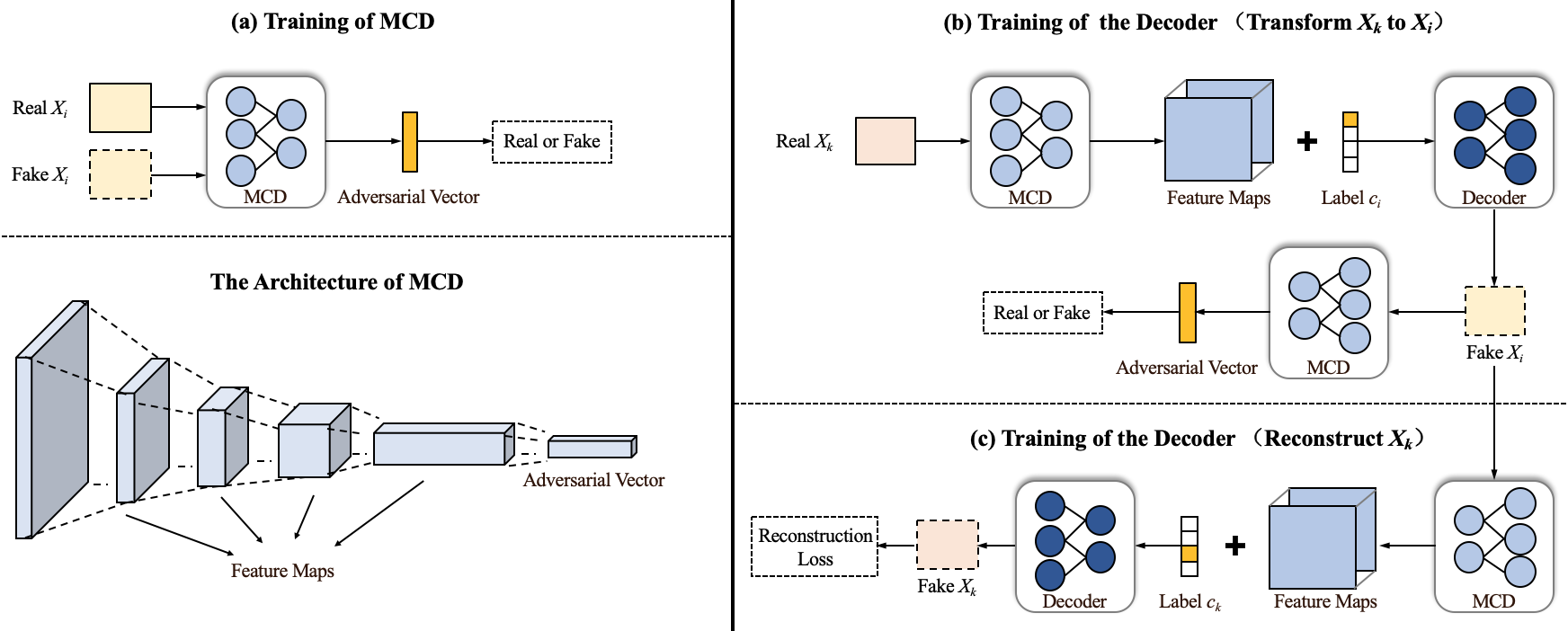}
\caption{The training process of $X_k\to X_i$. (a): $MCD$ learns to distinguish real $X_i$ and fake $X_i$. The input of $Decoder$ is the feature maps of $MCD$ and the label vector of $c_i$. The $i^{th}$ element of the adversarial vector $d$ is responsible for distinguishing real $X_i$ and fake $X_i$. We only optimize $MCD$ in this step. (b) and (c): $Decoder$ learns to translate the encoded information of $X_k$ to $X_i$ by giving the label vector of $c_i$. $Decoder$ tries to fool the $MCD$ to make fake $X_i$ indistinguishable with real $X_i$. Then $Decoder$ learns to translate the encoded information of fake $X_i$ to fake $X_k$ by giving the label vector $c_k$. The $L_1$ distance between real $X_k$ and fake $X_k$ is the reconstruction loss. We only optimize $Decoder$ in this step.}
\label{FigBig}
\end{figure*}

\subsection{CerfGAN}
CerfGAN implicitly matches the translation mappings
among multiple image domains. Fig. \ref{FigBig} shows the training process of translating $X_k\to X_i$. We denote the data distribution as $x^i\backsim P_{data}(X_i)$. In every iteration, MCD is first trained to encode the input and to distinguish if the input is real $X_i$. Then, the decoder is trained to fool MCD by feeding the encoded information of the input image and the target label vector. The objective function for training CerfGAN contains two terms, \emph{i.e.}, the multi-class adversarial loss for forcing the generated samples to be in the target domain, and the reconstruction loss for conserving the content and structure information from the input images to the generated images.

We then denote the feature maps of MCD as $M_{en}(x^k)$ by feeding the image sampled from the domain $X_k$. $DE(M_{en}(x^k), c_i)$ is the translated image by feeding $M_{en}(x^k)$ under the condition of the target label $c_i$.

\textbf{Multi-Class Adversarial Loss}. The multi-class adversarial loss is defined as:
\begin{equation}
\label{equMCDGANLoss}
    \begin{aligned}
    L_{adv}(i) = E_{x^i}\big[\log \big(M_{adv}(x^i)(i)\big)\big]\quad+ \ \qquad \ \quad \\E_{x^k,c_i}\Bigg[\log\Bigg(1-M_{adv}\Big(DE\big(M_{en}(x^k),c_i\big)\Big)(i)\Bigg)\Bigg],
    \end{aligned}
\end{equation}
where $M_{adv}(\cdot)(i)$ is the $i^{th}$ element of the adversarial vector as shown in Fig. \ref{Fig.Brief}. 

\textbf{Reconstruction Loss}. Minimizing the adversarial loss enables the decoder to output realistic images. However, the content and structure information of the input image cannot be conserved to the output. To address this problem, the reconstruction loss is applied to the decoder. 
\begin{equation}
\label{equAllRecLoss}
    \begin{aligned}
    L_{rec} = \qquad \ \qquad \ \qquad \ \qquad \ \qquad \ \qquad \ \qquad \ \qquad \ \quad \\E_{x^k,c_i,c_k}\Bigg[||x^k - DE\Bigg(M_{en}\Big(DE\big(M_{en}(x^k), c_i\big)\Big),c_k\Bigg)||_1\Bigg],
    \end{aligned}
\end{equation}
where $M_{en}\Big(DE\big(M_{en}(x^k), c_i\big)\Big)$ is the encoded information of the translated image. $DE$ translates it back to the image of the domain $X_k$ as the reconstructed image. The $L_1$ distance between the initial input image and the reconstructed image is defined as the reconstruction loss. 

\textbf{Full Objective Function}. The full objective function of CerfGAN is written as:
\begin{equation}
\label{equAllDLoss}
    L_M(i) = -L_{adv}(i),
\end{equation}
\begin{equation}
\label{equAllGLoss}
    L_{DE} = L_{adv}(i) + \lambda _{rec}L_{rec},
\end{equation}
where $i$ denotes that the decoder translates the input to the $i^{th}$ domain. $\lambda_{rec}$ is the hyper-parameter that controls the influence of the reconstruction loss to the full objective function. $\lambda_{rec}$ is 100 in all our experiments.

\section{Implementation}
\textbf{Network Architecture}. The network of CerfGAN is similar to U-Net \cite{ronneberger2015u}. But the principle is different. In terms of U-Net, there are skips between the encoder and the decoder to stabilize the training of the model, where gradients flow in these skips. However, in CerfGAN, we only use the feature maps of MCD as the input of the decoder. When training the decoder, MCD is fixed \footnote{In our supplementary material, we discuss how the quality of results is affected by using different layers of the feature maps as the input.}. In CerfGAN, the input images are first down-sampled by 6 stride-2 convolutions for 128$\times$128 resolution. The feature maps and the target label vectors are regarded as the input of the decoder that consists of 6 fractionally strided convolutions with stride 1/2. Then, one stride-2 convolution layer processes the encoded vector of the input image to the adversarial vector.
\begin{figure}[!htb]
\centering
\includegraphics[width=0.45\textwidth]{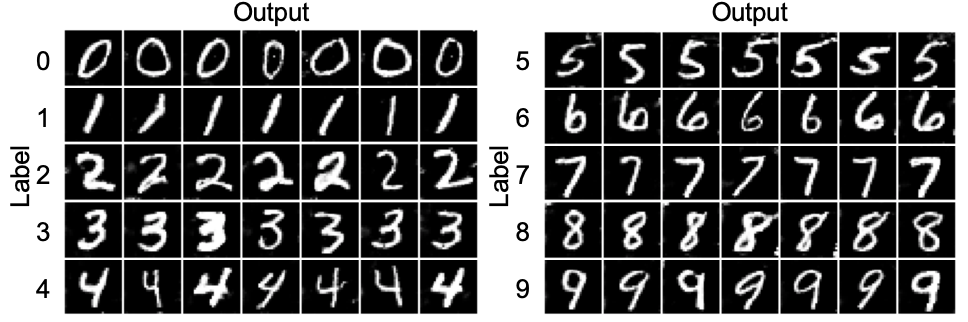}
\caption{The results of CGAN-MCD by giving different labels
and noises. The model is trained with MNIST dataset.}
\label{FigMNIST}
\end{figure}
\begin{figure}[!htb]
\centering
\includegraphics[width=0.45\textwidth]{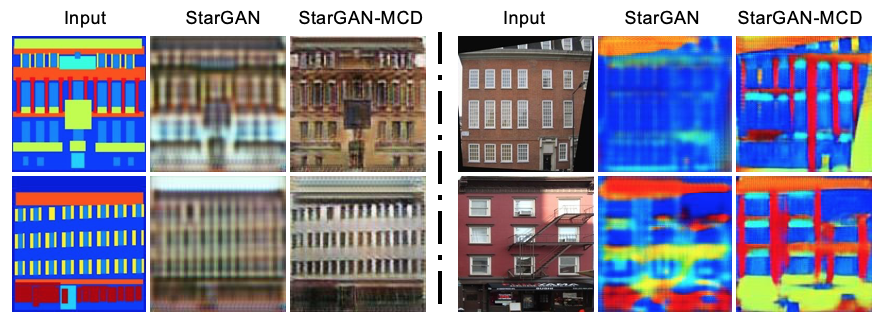}
\caption{Compare StarGAN and StarGAN-MCD in the task of facades labels $\rightleftharpoons$ photos.}
\label{FigStarMCD}
\end{figure}
Note that the one-hot label vector will be padded zeros to
let the dimension of the label vector be the same as the encoded image vector.

\textbf{Training Details}. Without loss of generality, we chose 8 well-known datasets to perform experiments. (1) We compare our model to StarGAN and CycleGAN by using 1,096 aerial maps and 1,096 aerial photographs from Google Maps \cite{isola2017image}, 2,975 cityscape semantic label images and 2,975 cityscape photos from the Cityscapes training set \cite{cordts2016cityscapes} and 400 architecture labels and 400 architecture photos from \cite{tylevcek2013spatial}. (2) We testify our model for one-to-one translation by using 1,273 summer photos and 854 winter photos from \cite{zhu2017unpaired}. (3) We testify our model for many-to-one and one-to-many translations (Style Transfer) by using 401 paintings of Van Gogh, 1,074 paintings of Monet, 584 paintings of Cezanne and 6,853 photographs from \cite{zhu2017unpaired}. (4) We testify our model in learning translation mappings from a very small dataset: 88 face photos and 88 face sketches from CUHK Students dataset \cite{wang2009face} \footnote{Note that the model is not trained to learn the translation mappings between irrelative domains like face photos and aerial photographs. Only a single model is trained to handle all of the tasks.}.

We randomly chose the pairs of domains to train the model in every epoch. The model was trained with $200\times N$ epochs, where $N$ is the number of domains. We use Adam Solver \cite{kingma2014adam} with $\beta 1=0.5$ and $\beta 2=0.999$ and set the batch size as 4 in all our experiments. The learning rate is 0.0002 in the first half of the training procedure. Then it linearly decays to 0 in the next half. Please see our supplementary material for more details of network architectures and training details.

\begin{figure}[!htb]
\centering
\includegraphics[width=0.45\textwidth]{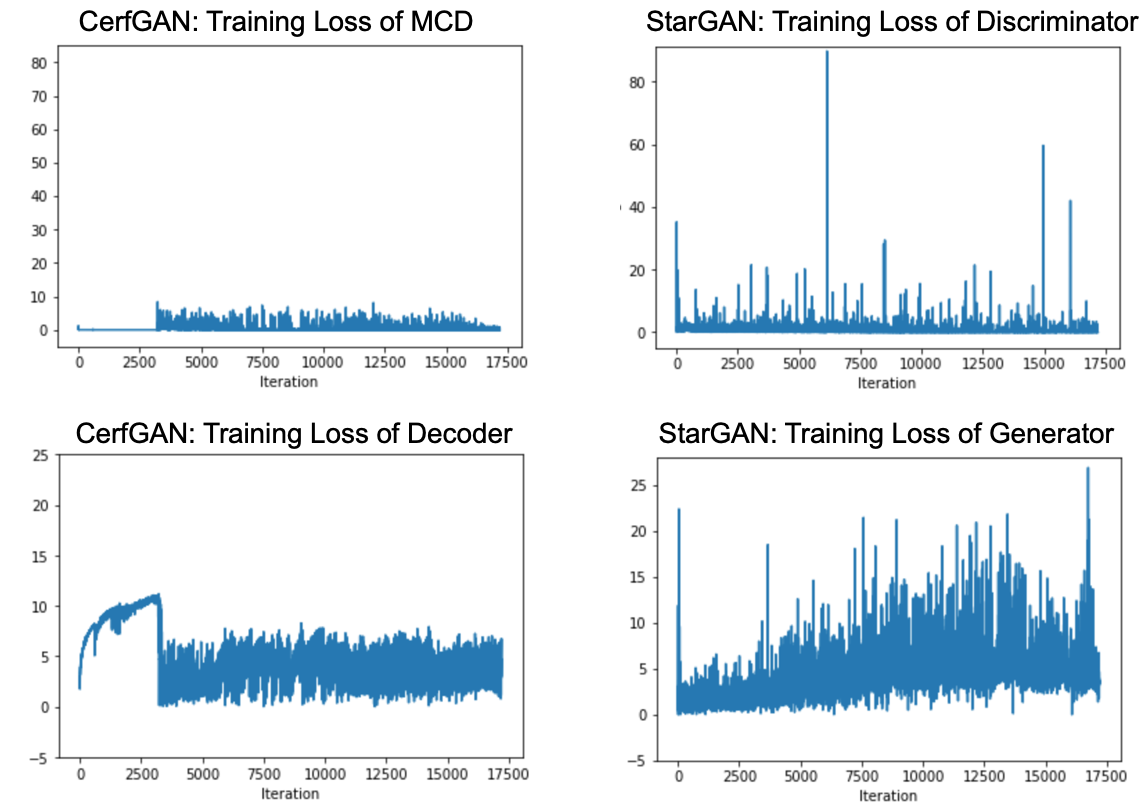}
\caption{Compare the training losses of CerfGAN and StarGAN.}
\label{FigGLoss}
\end{figure}

\section{Experiments}
In this section, we first do ablation studies to verify our discussions in Sec. \ref{CerfGAN}. Then we train CerfGAN to match the translation mappings between multiple domains. We compare our method with baseline models including pix2pix \cite{isola2017image}, CycleGAN \cite{zhu2017unpaired} and StarGAN \cite{choi2017stargan} on multiple translations tasks. Then, we demonstrate the superiority of our method by analyzing the results generated by our model, and by comparing the number of parameters with baseline models. Qualitative and quantitative comparisons are illustrated in Sec. \ref{Experimental Results}.

\subsection{Ablation Studies}
\label{SecLoss}
\textbf{Conditional Image Generation}. We verify the effectiveness of MCD by performing experiments on MNIST dataset. The network is based on CGAN \cite{mirza2014conditional}. We slightly change the binary discriminator of CGAN to MCD and use the label to choose the dimension of the adversarial vector, termed CGAN-MCD. As shown in Fig. \ref{FigMNIST}, CGAN-MCD successfully generates desired images by inputting different labels, which testifies the conclusion that training a decoder
by MCD can be formed as a new type of conditional GAN.

\begin{figure*}[!htb]
\centering
\includegraphics[width=1\textwidth]{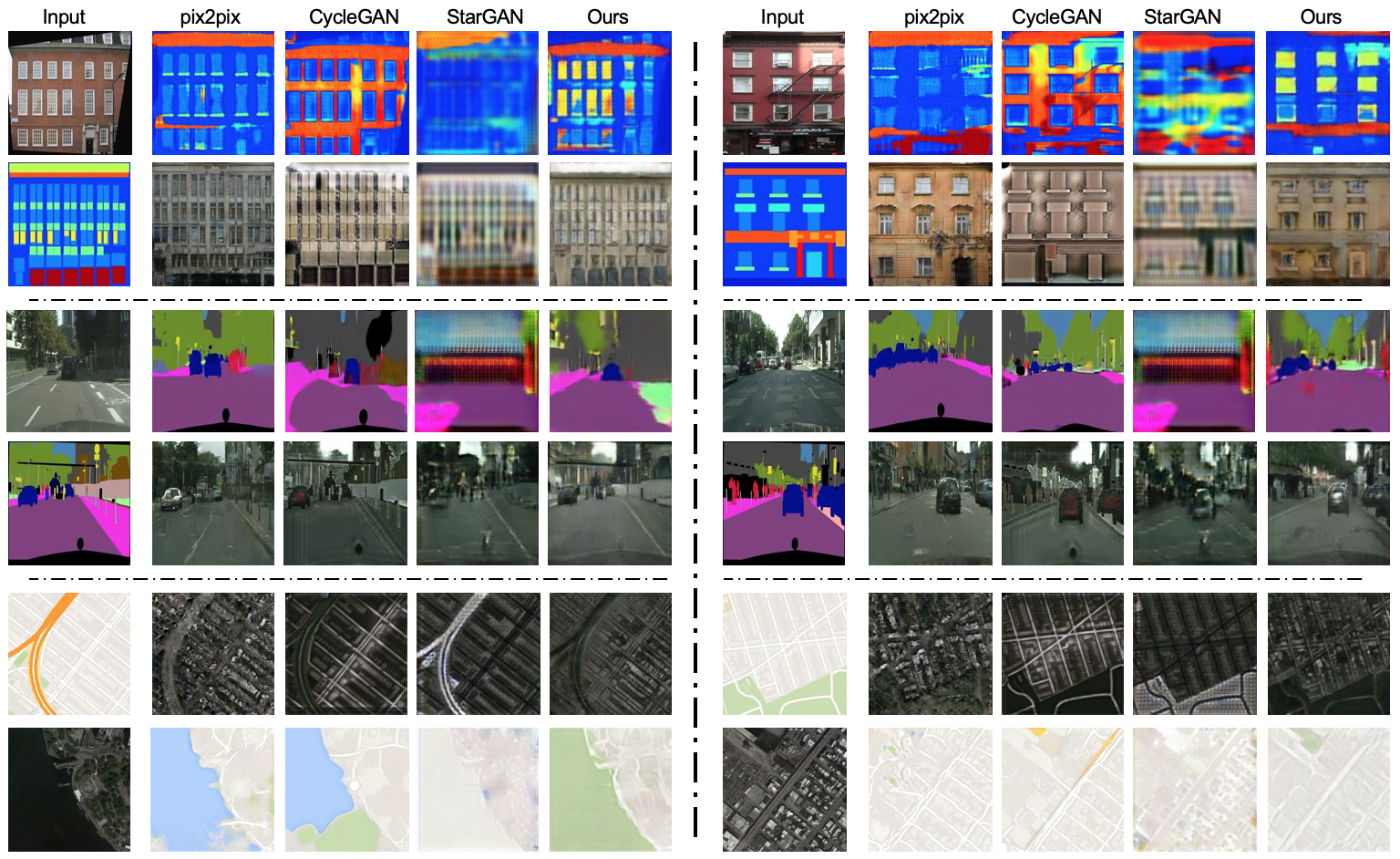}
\caption{The comparisons between our model with the baseline models. From up to down: facade labels $\rightleftharpoons$ photos, cityscape labels $\rightleftharpoons$ photos and aerial maps $\rightleftharpoons$ photos.}
\label{FigResultCom}
\end{figure*}

\textbf{StarGAN-MCD}. To testify that MCD can solve the problems of StarGAN, we delete the auxiliary classifier of StarGAN and change the discriminator of StarGAN to MCD, termed StarGAN-MCD. In Fig. \ref{FigStarMCD}, it is obvious that the original StarGAN has poor performance in the task of facade labels $\rightleftharpoons$ photos. Instead, training the generator by MCD performs much better than the original StarGAN in this task. Visually, the results of StarGAN-MCD contain much more details and have more style information of the target domain.

\textbf{CerfGAN: Stabilize the Training Process}. To demonstrate that CerfGAN is more stable to train, we plot the loss in Fig. \ref{FigGLoss}. The dynamic ranges of the loss of CerfGAN are considerably narrower than that of StarGAN. And the loss of CerfGAN does not have very large values, which means there is almost no gradient exploding problem.

\subsection{Baseline Models}
We compare our model to pix2pix \cite{isola2017image} and CycleGAN \cite{zhu2017unpaired} to demonstrate that our method is able to translate images between considerably different domains. We also compare our model to StarGAN to illustrate that our framework is general for multi-domain image-to-image translation.

\textbf{pix2pix} has a similar architecture to our framework except the encoder. pix2pix optimizes the reconstruction loss $||y-G_{XY}(x)||_1$ and the adversarial loss to train the model on paired data, which is supposed to have the upper-bound quality.

\textbf{CycleGAN} trains two models to find the mappings between two domains $X$ and $Y$. It applies the cycle-consistency loss $||x-G_{YX}(G_{XY}(x))||_1$ and $||y-G_{XY}(G_{YX}(y))||_1$ to force the generated images to keep the content and structure information of the inputs. 

\textbf{StarGAN} utilizes one discriminator, one encoder and one decoder to learn the mappings between multiple domains. By jointly minimizing the classification loss, the reconstruction loss and the adversarial loss, the outputs of the generator are constrained into the result domains. 

\subsection{Experimental Results}
\label{Experimental Results}
We first train pix2pix and CycleGAN respectively by
feeding images from pairs of domains. Then, we train our
proposed model and StarGAN as unified models on multiple
domains, where a single model handles all tasks. All models are trained to translates 128$\times$128 images.

\begin{figure*}[!htb]
\centering
\includegraphics[width=1\textwidth]{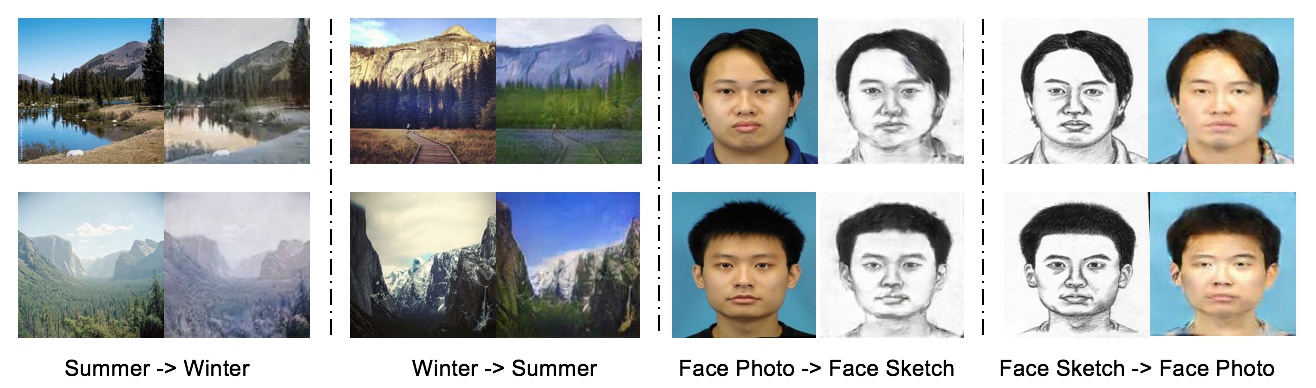}
\caption{Results of Season Transfer and Face Hallucination}
\label{FigResult1}
\end{figure*}

\begin{figure*}[!htb]
\centering
\includegraphics[width=1\textwidth]{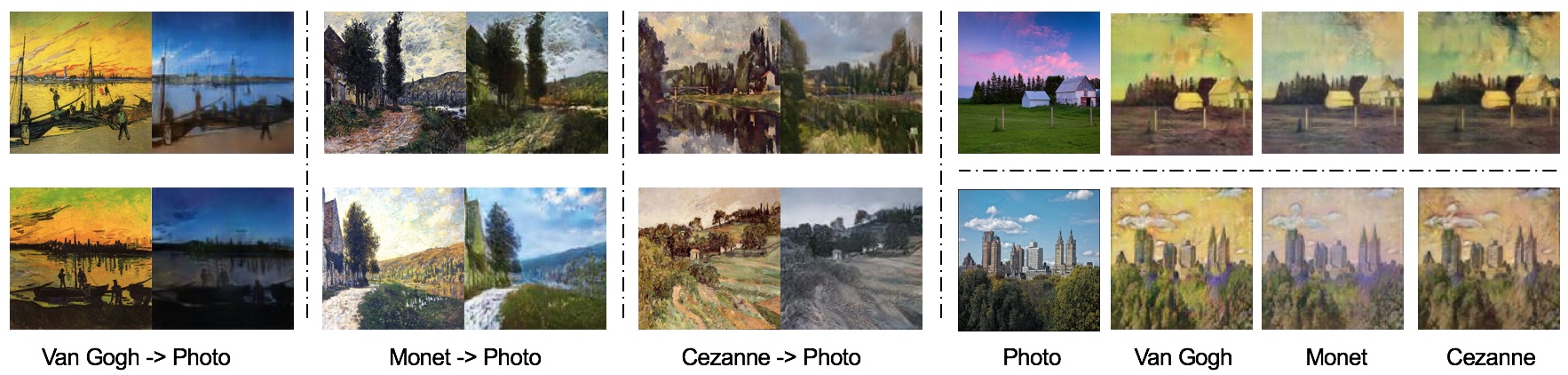}
\caption{Results of Style Transfer}
\label{FigResult2}
\end{figure*}

\textbf{Qualitative Evaluation}
In Fig. \ref{FigResultCom}, we compare our results with pix2pix, CycleGAN and StarGAN. Qualitatively, our unsupervised model is comparable with a limited gap
to the supervised method, pix2pix. Comparing to Cycle-
GAN, our method outputs results with similar visual quality. Note that facade labels $\rightleftharpoons$ photos, cityscape labels $\rightleftharpoons$ photos and aerial maps $\rightleftharpoons$ photos are considerably different. As a result, StarGAN fails to learn the translation mappings
in these tasks. Comparing to StarGAN, we successfully
translate all tasks by the unified model.

\begin{table}[!htb]\footnotesize
      \centering
      \caption{Quantitative Evaluation}
      \scalebox{0.9}{
       \begin{tabular}{cccc}
         \toprule
         Method  &  Label $\to$ Photo  &  Photo $\to$ Label  &  Average  \\
         \midrule
         CycleGAN   &   33.67\%   &   \textbf{51.53}\%   
         &  42.6\% \\
         StarGAN   &   9.18\%   &   2.55\%   
         &  5.87\% \\
         CerfGAN   &   \textbf{57.15}\%   &   45.92\%   
         &  \textbf{51.53}\% \\
         \bottomrule
       \end{tabular}
       } 
    \label{TableQuan}
\end{table}
\begin{table}[!htb]\footnotesize
      \centering
      \caption{Parameters and Training Time}
      \scalebox{0.9}{
       \begin{tabular}{ccc}
         \toprule
         Method  &   Parameters  & Training Time (on CelebA)\\
         \midrule
         CycleGAN  &  52.6M$\times$16 & -\\
         StarGAN   &  53.2M$\times$1 & 24 hours\\
         CerfGAN   &   \textbf{34.1}M$\times$1 & \textbf{19} hours\\
         \bottomrule
       \end{tabular}
       } 
    \label{TableTime}
\end{table}

CerfGAN is trained to learn the mappings between 16 domains. Fig. \ref{FigResult1} and Fig. \ref{FigResult2} show the results of season transfer, face hallucination and style transfer. Note that there are 8 datasets including 16 domains with 3 domains sharing the same data, \emph{i.e.}, Van Gogh $\rightleftharpoons$ Photos, Monet $\rightleftharpoons$ Photos and Cezanne $\rightleftharpoons$ Photos use the same set of photos as the training data. On the other hand, although, there are only 88 pairs of training images in CUHK Students dataset. our model successfully translates images of this task without any specific setting, which testifies the effectiveness and robustness of CerfGAN.

\textbf{Quantitative Evaluation}. For quantitative evaluations, we perform a user study to assess the translation quality between facade labels and photos. The users were instructed to choose the best generated image of the input based on the perceptual realism and the consistency of content and structure. In each question, we gave a pair of images of the translation task as an example to the user. One question had three options, \emph{i.e.}, three images generated by CycleGAN, StarGAN and CerfGAN (order randomly). We asked the user which option was the best translated image by giving the input. Moreover, there were a few logical questions for validating human effort. As shown in Tab. \ref{TableQuan}, both our model and CycleGAN obtain around half votes for the best generated images, while few users voted for StarGAN, which indicates that the quality of the generated images of our model is much better than StarGAN and is close to CycleGAN. 

Since MCD also works as an encoder in CerfGAN, as shown in Tab. \ref{TableTime}, the number of parameters of CerfGAN is much less than CycleGAN and StarGAN. Specifically, CerfGAN is more \emph{compact}, which has only 34.1M parameters that are 64.1\% of StarGAN and 0.041\% of CycleGAN. The training time of CerfGAN is around 79.17\% of StarGAN, referring the merit of \emph{fast} of CerfGAN.

\section{Conclusion and Future Work}
In this paper, CerfGAN as a novel model is proposed for
multi-domain image-to-image translation. In particular, we
first analyze that StarGAN is unable to solve general multidomain translation problems and unstable to train. To address these problems, we propose the multi-class discriminator (MCD), which also plays a role of the encoder in CerfGAN. Experiments demonstrate that our proposed model is more stable to train and works better in multi-domain translations with high domain shifts. Qualitative and quantitative results testify the four merits of CerfGAN, \emph{i.e.}, \emph{compact} (only 64.1\% parameters of StarGAN), \emph{effective} and \emph{robust} (a single model handles 8 datasets), \emph{fast} (around 79.17\% training time of StarGAN).

\textbf{Acknowledgments}: We thank Miss Qisi Zhang for helpful discussions and her work on proof-reading. This work is supported by the National Key R\&D Program (No.2017YFC0113000, and No.2016YFB1001503), Nature Science Foundation of China (No.U1705262, No.61772443, and No.61572410), Post Doctoral Innovative Talent Support Program under Grant BX201600094, China Post-Doctoral Science Foundation under Grant 2017M612134, Scientific Research Project of National Language Committee of China (Grant No. YB135-49), and Nature Science Foundation of Fujian Province, China (No. 2017J01125 and No. 2018J01106).


{\small
\bibliographystyle{ieee}
\bibliography{egbib}
}

\newpage \ \newpage
\section{Supplementary Material}
\subsection{WGAN-GP Loss}
\begin{figure}[!htb]
\centering
\includegraphics[width=0.45\textwidth]{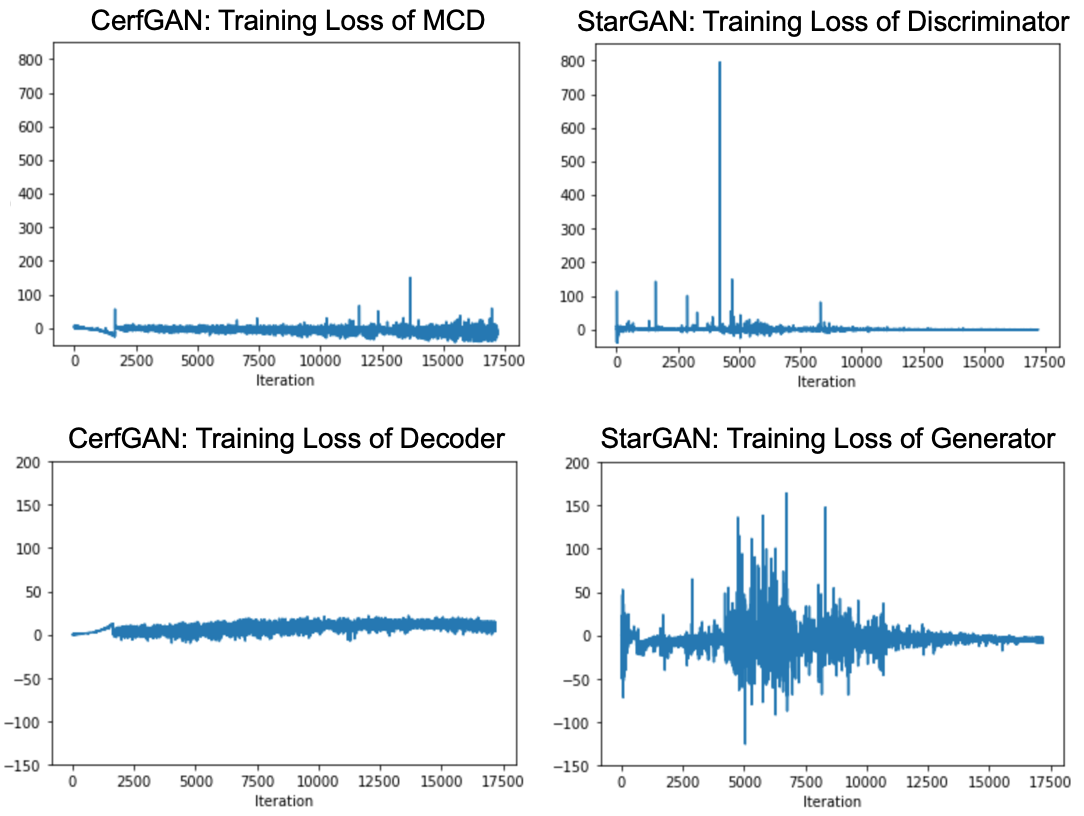}
\caption{Use WGAN-GP loss to train the StarGAN and CerfGAN. Comparison of the training loss of CerfGAN and StarGAN.}
\label{WGLoss}
\end{figure}

In the paper, we used initial GAN loss to train StarGAN and CerfGAN. As StarGAN used WGAN-GP \cite{gulrajani2017improved} loss to train the model, we also performed experiments by using WGAN-GP loss to train CerfGAN. From Fig. \ref{WGLoss}, we can also have the conclusion that the dynamic ranges of the loss of CerfGAN are considerably narrower than the loss of StarGAN (EDD architecture). And the loss of CerfGAN has few very large values, which means there is almost no gradient exploding problem. Meanwhile, the loss of the discriminator of StarGAN does not fluctuate at all in the second half of the training process. The reason is that the gradient vanishes a lot in this stage. On the other hand, we find that when we use WGAN-GP loss to train CerfGAN, the results have a better visual quality but are less translated into the desired domain. So our model generating all images in this paper is trained by initial GAN loss.

\subsection{Network Architecture and Training Details}
\label{Appendix Network}
The network architectures for 128$\times$128-resolution translation are listed below in Tab. \ref{TableArchiD} and Tab. \ref{TableArchiG}. We use batch normalization \cite{ioffe2015batch} in all layers of MCD except the first layer and the output layer. Leaky ReLU \cite{maas2013rectifier} with a negative scope of 0.2 is used in all layers of MCD except the output layer. For the architecture of the decoder, we use batch normalization and ReLU \cite{nair2010rectified} in all layers except the last layer, where we use Tanh function. The notations in the tables:
N: the number of domains;
O: the number of output channels;
K: the kernel size;
S: the stride size;
P: the padding size;
BN: batch normalization;

We randomly crop all images to 128$\times$128. Then we randomly mirror and jitter the images in the pre-processing step. All networks are trained from scratch. Weights are initialized from a Gaussian distribution with mean 0 and standard deviation 0.02.  

\begin{table*}[!htb]\footnotesize
      \centering
    \caption{The architecture of MCD}
      \begin{tabular}{c|r@{$\to$}l|c}
        \hline
        Part &  Input Shape & Output Shape   & Layer Information\\
        \hline
        \multirow{7}*{Down-Sampling ($M_{en}$)} & (3,128,128)&(64,128,128)   &   CONV-(O:64,K:7x7,S:1,P:3), Leaky ReLU\\
        ~ & (64,128,128)    &(64,64,64)	&   CONV-(O:64,K:4x4,S:2,P:1), BN, Leaky ReLU\\
        ~ & (64,64,64)  & (128,32,32)	&   CONV-(O:128,K:4x4,S:2,P:1), BN, Leaky ReLU\\
        ~ & (128,32,32)& (256,16,16)	&   CONV-(O:256,K:4x4,S:2,P:1), BN, Leaky ReLU\\
        ~ & (256,16,16)&(512,8,8)	    &   CONV-(O:512,K:4x4,S:2,P:1), BN, Leaky ReLU\\
        ~ & (512,8,8)   &(512,4,4)	    &   CONV-(O:512,K:4x4,S:2,P:1), BN, Leaky ReLU\\
        ~ & (512,4,4)& (512,2,2)	    &   CONV-(O:512,K:4x4,S:2,P:1), BN, Leaky ReLU\\
        \hline
        Output Layer ($M_{adv}$) & (512,2,2)&(N,1,1) & CONV-(O:N,K:4x4,S:2,P:1) \\
        \hline
      \end{tabular}
    \label{TableArchiD}
\end{table*}

\begin{table*}[!htb]\footnotesize
      \centering
    \caption{The architecture of the decoder}
      \begin{tabular}{c|r@{$\to$}l|c}
        \hline
        Part &  Input Shape & Output Shape   & Layer Information\\
        \hline
        \multirow{7}*{Up-Sampling ($DE$)} & (1024,2,2)&(512,4,4)  &   DECONV-(O:512,K:4x4,S:2,P:1), BN, ReLU\\
        ~ & (1024,4,4)    &(512,8,8)	&   DECONV-(O:512,K:4x4,S:2,P:1), BN, ReLU\\
        ~ & (1024,8,8)  & (256,16,16)	&   DECONV-(O:256,K:4x4,S:2,P:1), BN, ReLU\\
        ~ & (512,16,16)& (128,32,32)	&   DECONV-(O:128,K:4x4,S:2,P:1), BN, ReLU\\
        ~ & (256,32,32)&(64,64,64)	    &   DECONV-(O:64,K:4x4,S:2,P:1), BN, ReLU\\
        ~ & (64,64,64)   &(64,128,128)	    &   DECONV-(O:64,K:4x4,S:2,P:1), BN, ReLU\\
        ~ & (64,128,128)& (3,128,128)	    &   DECONV-(O:3,K:7x7,S:1,P:3), Tanh\\
        \hline
      \end{tabular}
    \label{TableArchiG}
\end{table*}

\begin{figure}[!htb]
\centering
\includegraphics[width=0.45\textwidth]{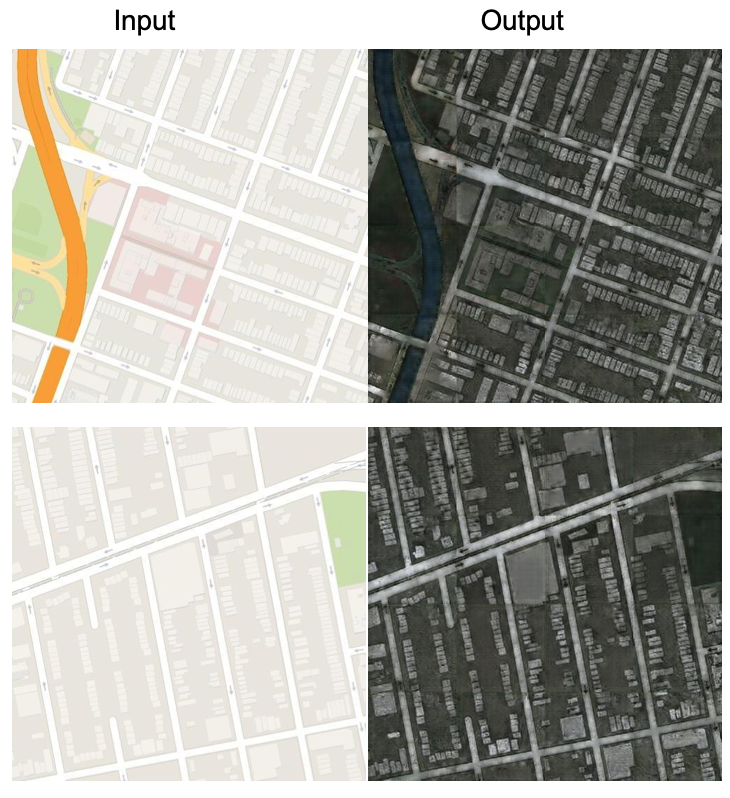}
\caption{The high-resolution images generated by CerfGAN.}
\label{FigResultsHR}
\end{figure}

\subsection{High-Resolution Image-to-Image Translation}
As we randomly crop the training data to 128$\times$128 to train the model, our model is able to generate high-resolution images (512$\times$512). We run our model convolutionally on the 512$\times$512 aerial maps at the test time. Fig. \ref{FigResultsHR} shows the high-resolution images generated by our model.

\subsection{Translations between Irrelative Domains}
\begin{figure*}[!htb]
\centering
\includegraphics[width=1\textwidth]{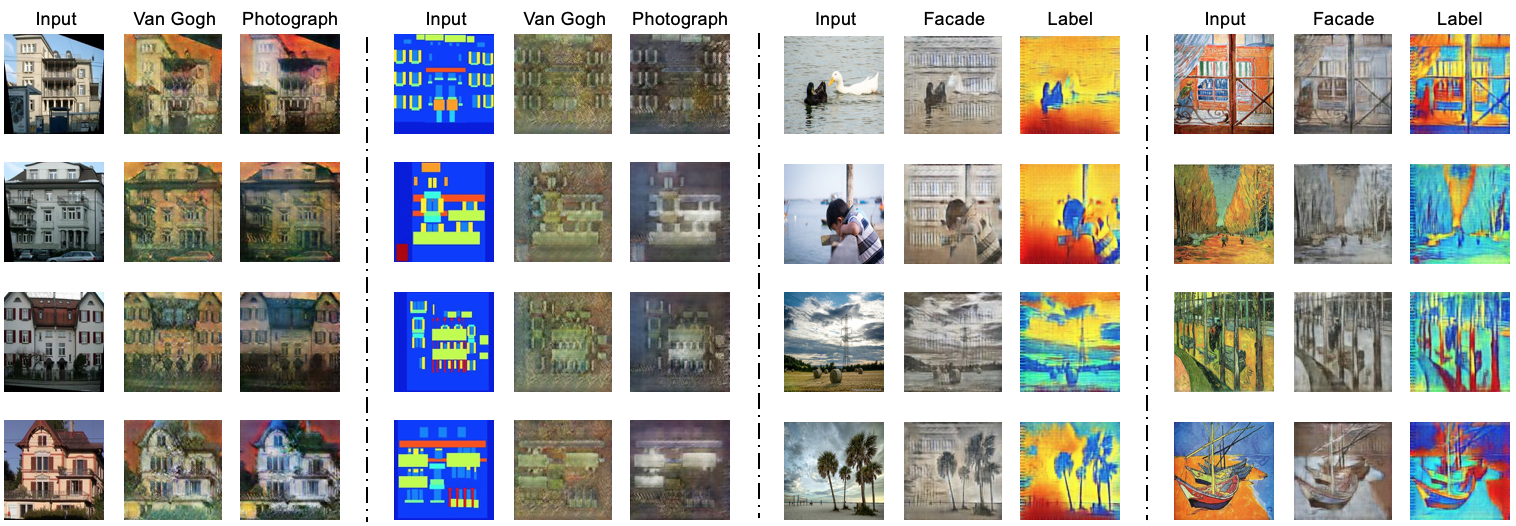}
\caption{Translations between irrelative domains.}
\label{FigVtoS}
\end{figure*}

Principally, CerfGAN is able to learn thousands of translation mappings between general image domains like CNNs having the ability of classifying thousands of classes. Without loss of generality, we chose 8 datasets to train CerfGAN. Although, we manually set CerfGAN not to learn the translation mappings between irrelative domains like facades photos and Van Gogh images. However, if we train CerfGAN with a massive number of datasets, transfer leaning methods are possible to be used to simplify the training process of CerfGAN, where some datasets may be irrelative, which will be our future work. So, it is also meaningful to try the translations between irrelative domains. We performed experiments to show the powerful generative ability of CerfGAN by training the model with irrelative datasets. As shown in Fig. \ref{FigVtoS}, CerfGAN is also able to learn translation mappings between Van Gogh, photos, facade labels and facade photos. 

\subsection{Different Layers as the Input of the Decoder}
\begin{figure*}[!htb]
\centering
\includegraphics[width=1\textwidth]{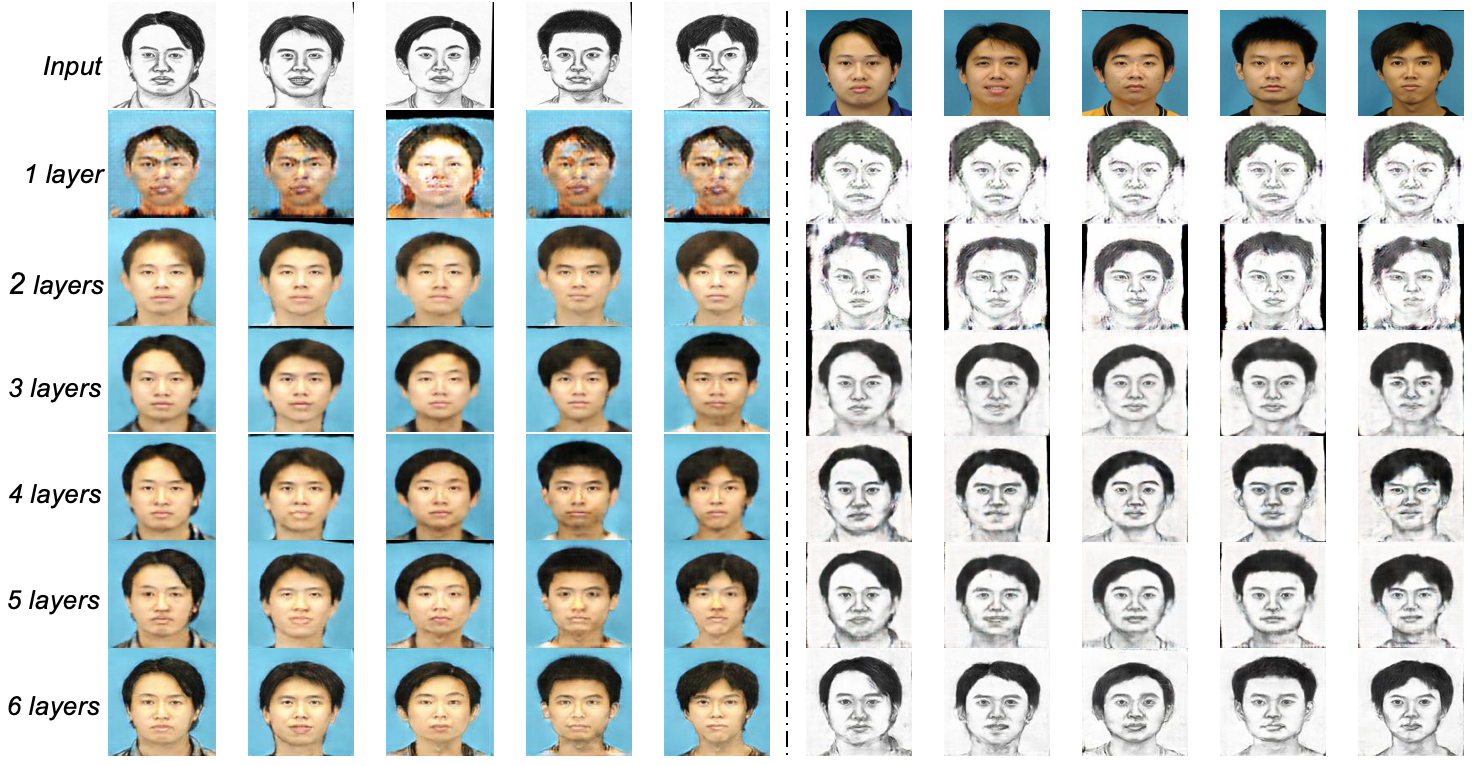}
\caption{Input different feature maps of MCD to the decoder. CUHK Photo $\rightleftharpoons$ Sketch.}
\label{UNet}
\end{figure*}

In terms of the architecture of CerfGAN, the input of the decoder is the feature maps of MCD. When we use different feature maps of MCD as the input, the outputs of the decoder are very different. As shown in Fig. \ref{UNet}, when the input is only the first layer of the feature maps, the decoder also translates the input image to the target domain, which is the purpose of MCD. But the output images have low visual quality and these images are very similar to each other. When we use more layers as the input, the quality increases and the output images keep more structure and content information of the input images.

\subsection{Face Attribute Modulation by CerfGAN}
\begin{figure*}[!htb]
\centering
\includegraphics[width=1\textwidth]{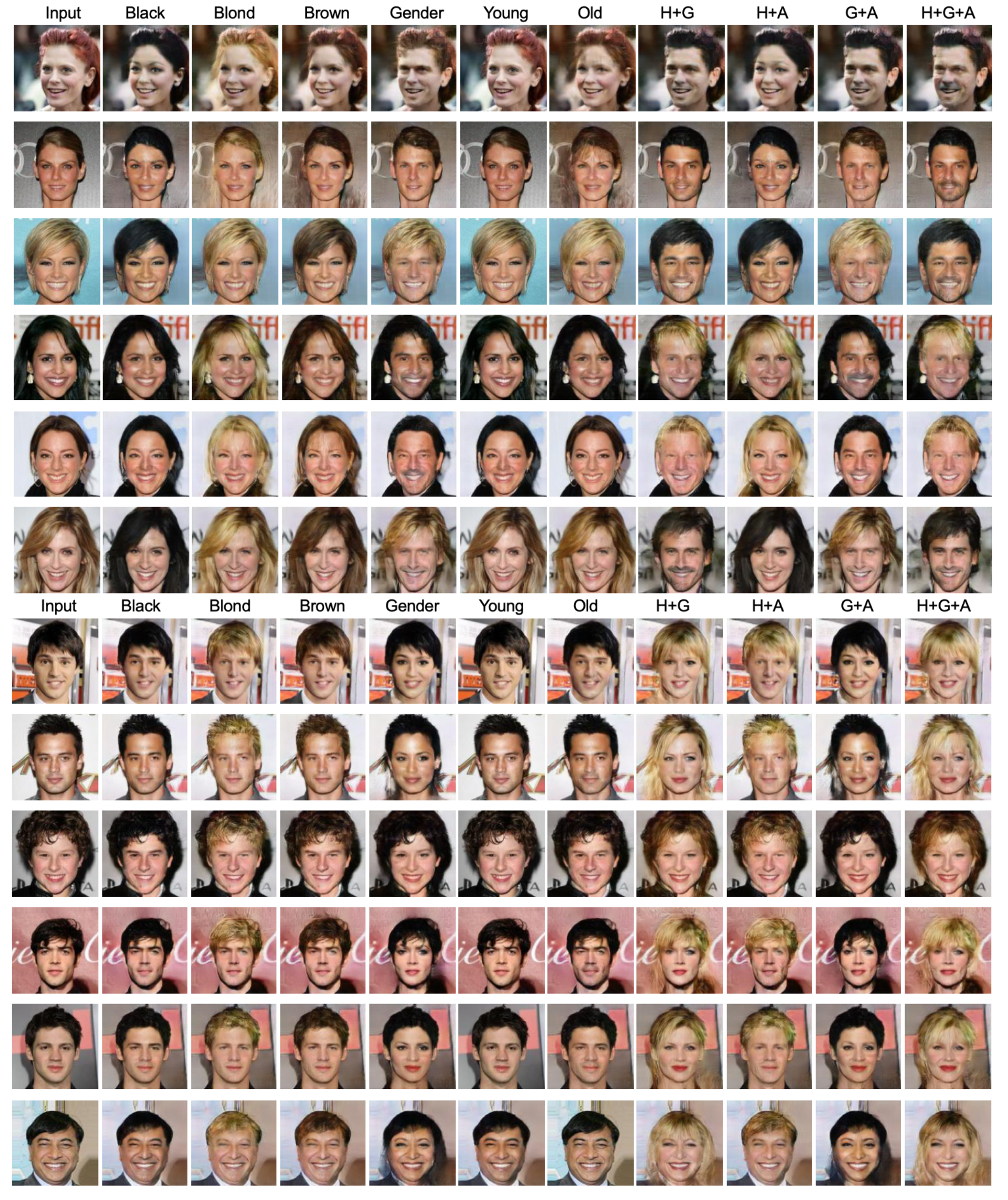}
\caption{Face Attribute Modulation. Single and multiple attribute transfer on CelebA. H+G: Hair Color + Gender. H+A: Hair Color + Old/Young. G+A: Gender + Old/Young. H+G+A: Hair Color + Male/Female + Old/Young.}
\label{Cel}
\end{figure*}

In principle, CerfGAN is a more general method for multi-domain image-to-image translation tasks than StarGAN. We analyzed that StarGAN is unsuitable for general tasks. However, StarGAN performs very well in face attribute modulation. To demonstrate the superiority of CerfGAN, we also performed experiments in face attributes modulation with CelebA dataset \cite{liu2015faceattributes}. In CelebA, there are 202,599 face images and each image has 40 labeled binary attributes. When training CerfGAN with CelebA dataset, each image is from multiple domains. For example, the face in a image can have such a label, (black hairs , male, young) , which means the image is from three domains. In this case, the output of MCD in CerfGAN is multiple dimensions of the adversarial vector. Similar to StarGAN, CerfGAN can also do single and multiple attributes translations. In Fig. \ref{Cel}, the results show that CerfGAN is able to learn translation mappings when slight changes are required.

\subsection{More Results}
\begin{figure*}[!htb]
\centering
\includegraphics[width=1\textwidth]{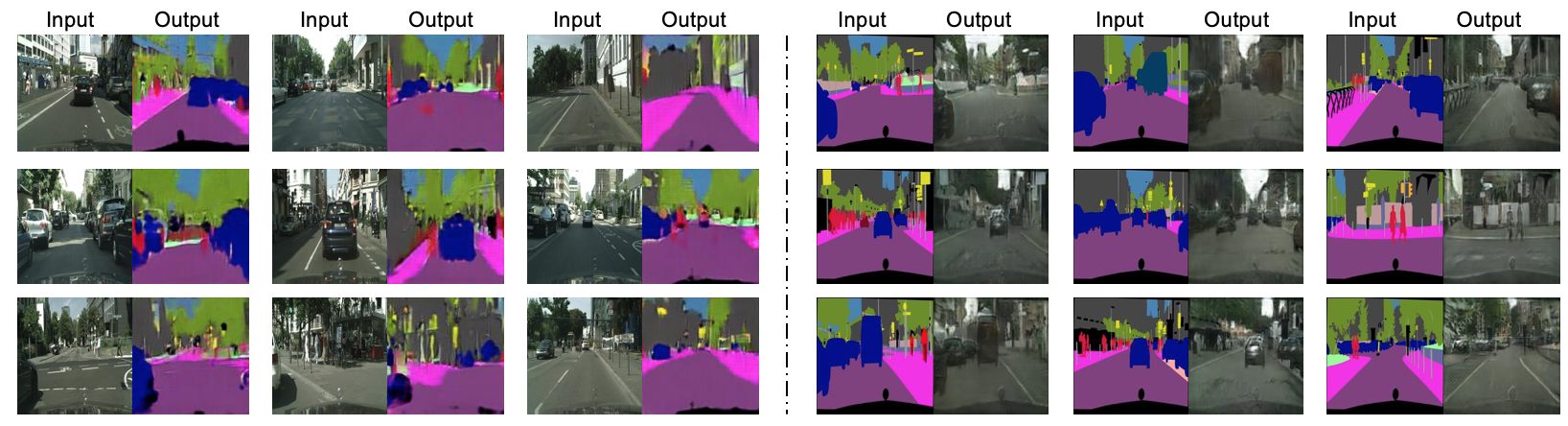}
\caption{Cityscapes Labels $\rightleftharpoons$ Photos. Left: Photos to Labels. Right: Labels to Photos.}
\label{MR1}
\end{figure*}

\begin{figure*}[!htb]
\centering
\includegraphics[width=1\textwidth]{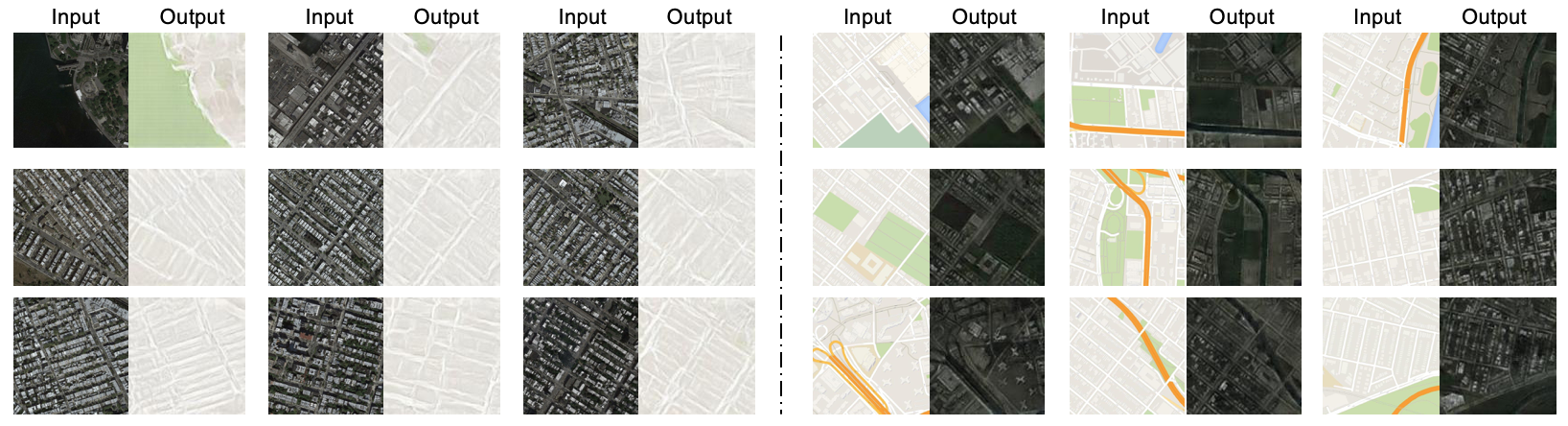}
\caption{Maps $\rightleftharpoons$ Photos. Left: Maps to Photos. Right: Photos to Maps.}
\label{MR2}
\end{figure*}

\begin{figure*}[!htb]
\centering
\includegraphics[width=1\textwidth]{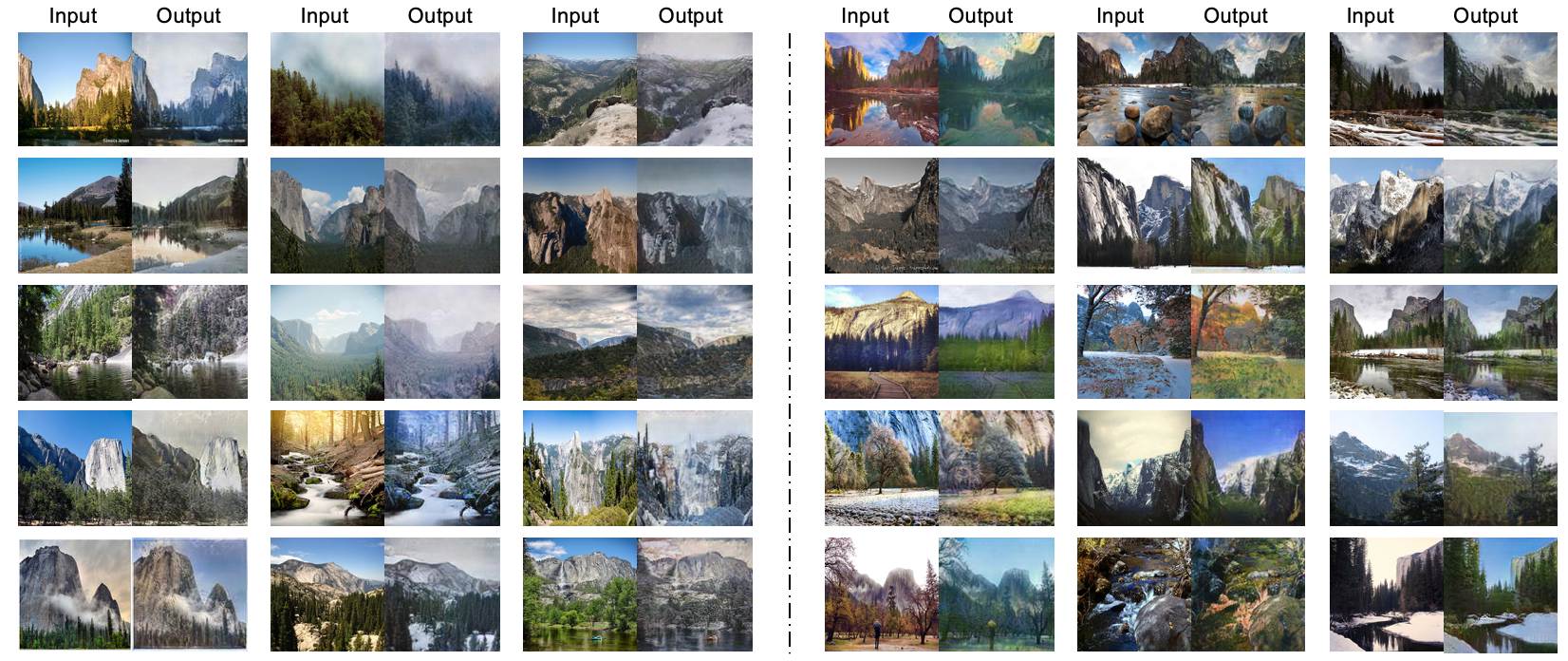}
\caption{Summer $\rightleftharpoons$ Winter. Left: Summer to Winter. Right: Winter to Summer.}
\label{MR3}
\end{figure*}

\begin{figure*}[!htb]
\centering
\includegraphics[width=1\textwidth]{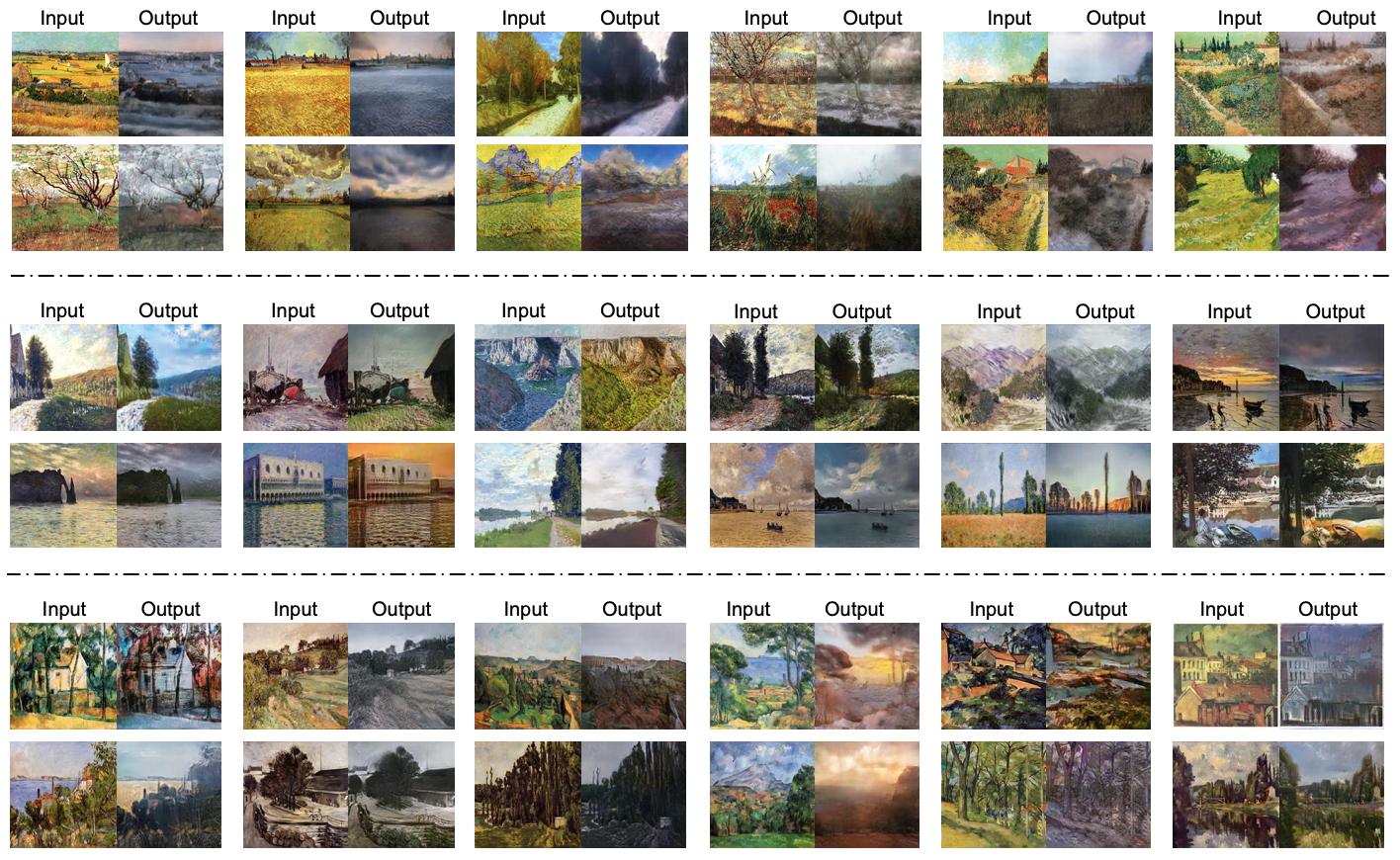}
\caption{Style Transfer. Up: Van Gogh to Photos. Middle: Monet to Photos. Bottom: Cezanne to Photos.}
\label{StyleMR1}
\end{figure*}

\begin{figure*}[!htb]
\centering
\includegraphics[width=1\textwidth]{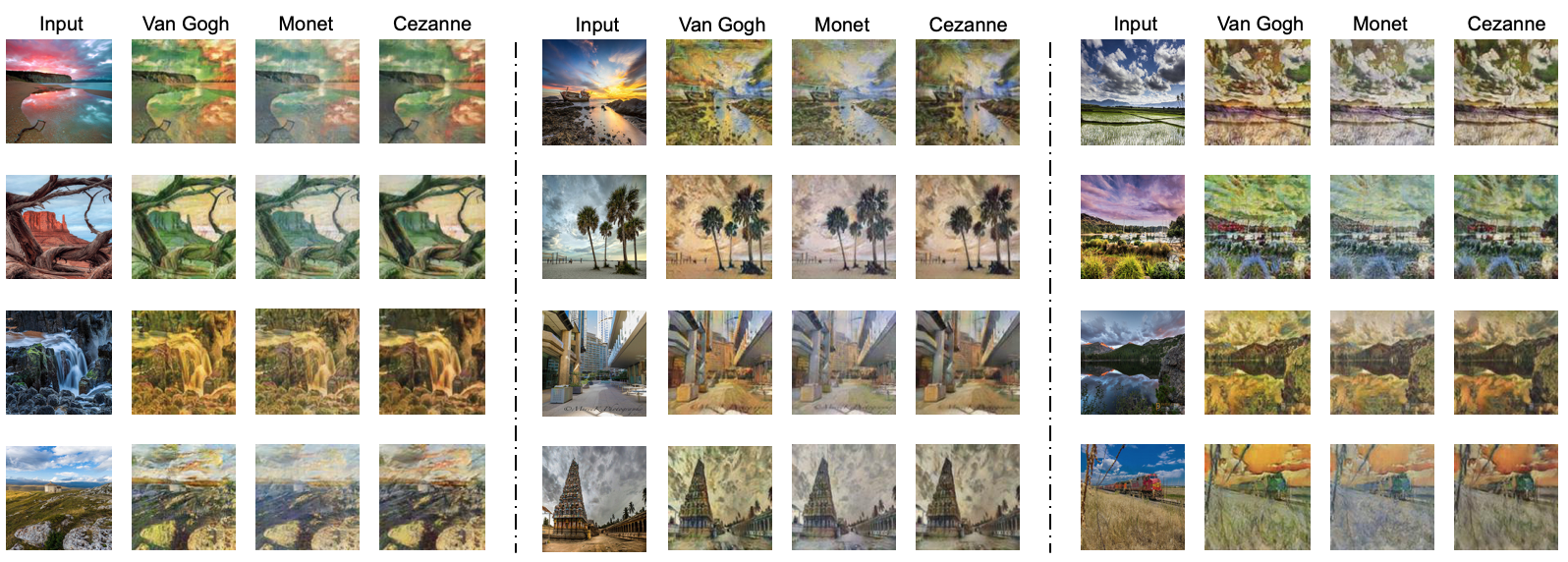}
\caption{Style Transfer. Photos to Van Gogh, Monet and Cezanne.}
\label{StyleMR2}
\end{figure*}

All results shown in Fig. \ref{MR1}, Fig. \ref{MR2}, Fig. \ref{MR3}, Fig. \ref{StyleMR1} and Fig. \ref{StyleMR2} are generated by a single model. 

\end{document}